\documentclass[a4paper, 10pt, conference]{ieeeconf}      
\usepackage{FG2023}
\usepackage{graphicx}
\usepackage{amsmath}
\usepackage{amsfonts}
\usepackage[noadjust, sort, compress]{cite}
\newcommand*{\Scale}[2][4]{\scalebox{#1}{$#2$}}%
\newcommand{\cut}[1]{}
\FGfinalcopy 

\IEEEoverridecommandlockouts                              
\def\FGPaperID{37} 

\title{\LARGE \bf SS-VAERR: Self-Supervised Apparent Emotional Reaction \\
Recognition from Video
}

\author{\parbox{16cm}{\centering
    {\large Marija Jegorova$^1$, Stavros Petridis$^{1,2}$, and Maja Pantic$^{1,2}$}\\
    {\normalsize
    $^1$ Meta Reality Labs, London, United Kingdom\\
    $^2$ Department of Computing, Imperial College London, United Kingdom}}
    \thanks{This work has been supported by Meta Reality Labs. With the exception of training the pre-text architectures on LRS3 dataset, which has been conducted on the servers of Imperial College London.}
}

\begin{document}

\ifFGfinal
\thispagestyle{empty}
\pagestyle{empty}
\else
\author{Marija Jegorova, Stavros Petridis~\IEEEmembership{Member,~IEEE}, Maja Pantic~\IEEEmembership{Fellow~Member,~IEEE}}
\pagestyle{plain}
\fi
\maketitle

\bstctlcite{IEEEexample:BSTcontrol} 

\begin{abstract}

This work focuses on the apparent emotional reaction recognition (AERR) from the video-only input, conducted in a self-supervised fashion. The network is first pre-trained on different self-supervised pretext tasks and later fine-tuned on the downstream target task. Self-supervised learning facilitates the use of pre-trained architectures and larger datasets that might be deemed unfit for the target task and yet might be useful to learn informative representations and hence provide useful initializations for further fine-tuning on smaller more suitable data. Our presented contribution is two-fold: (1) an analysis of different state-of-the-art (SOTA) pretext tasks for the video-only apparent emotional reaction recognition architecture, and (2) an analysis of various combinations of the regression and classification losses that are likely to improve the performance further. Together these two contributions result in the current state-of-the-art performance for the video-only spontaneous apparent emotional reaction recognition with continuous annotations.

\end{abstract}



\section{Introduction}\label{sec:intro}
Apparent emotional reaction recognition (AERR) is a broadly applicable branch of computer vision.
In this paper we are going to focus on specifically the video-only domain for AERR for several reasons. First, the audio stream is not always available\cut{ in the real life, for example in most CCTV cases. Second, even if it is available, correctly assigning a voice to a face becomes a struggle in case of multiple participants in the video. Finally, real-life noisy environments can represent certain challenges to audio-emotion recognition.}\textcolor{black}{, and not every apparent emotional reaction is accompanied by a sound. Second, in audio-visual domain active speaker detection is a whole new problem in case of multiple speakers in the video. Finally, generalising to noisy environments can represent certain challenges for audio}. Hence it would be useful to explore the efficient AERR restricted solely to the video modality for the sake of prediction robustness and broader applicability.

Further, this work explores predicting the continuous emotion characteristics - arousal and valence (in this paper we call these \textit{continuous emotions}) instead of more traditional AERR that is concerned with classifying the \textit{categorical emotions} (sadness, fear, surprise, etc). The reason being that the categorical emotion theory is limited in its ability to express subtle and disparate emotions \cite{ArousalValenceOverDiscreteEmotions}.

Current state-of-the-art for video-only AERR are \cite{AffectiveProcessess} and \cite{RECOLA_VideoOnly}. First presents a model based on probabilistic modeling of the temporal context, presenting compelling results on SEWA dataset \cite{SEWA}. A somewhat comparable performance is achieved by \cite{video-only-sewa-Kossaifi2020FactorizedHC}, using spatio-temporal higher-order convolutional neural net. Secondly, for RECOLA dataset \cite{RECOLA} the current SOTA is TS-SATCN \cite{RECOLA_VideoOnly}, a two-stage spatio-temporal attention temporal convolution network. The only additional video-only AERR method to be found is Visual ResNet-50 presented in \cite{RECOLA_VisualResNet50}, also evaluated on RECOLA. 

\cut{It is a frequent issue with real life applications, however, that there is little to none annotated data available for the specific tasks. That could potentially mean a limited performance and generalization on real life applications.}
Shortage of annotated data for specific tasks and domains often represents a challenge.
This can be addressed from several angles, e.g. transfer or semi-supervised learning and self-supervised learning (SSL). We focus on the SSL approach, that can use labelled and unlabelled data within the same model. It relies on the pretext training to leverage the additional data, and then serve as an initialization to the downstream training, solving target tasks. 

The works that contributed to the SSL paradigm for facial data in adjacent domains are \cite{Bulat_PreTrainingStratediesForFaceRepresentationLearning} and \cite{SelfSupervised_SpatioTemporal_contrastiveFER}. One presented a SSL framework for a number of tasks, including the AERR from images, providing results on AffectNet, large-scale facial expression image database \cite{AffectNet}, and is the SOTA for self-supervised AERR on images \cite{Bulat_PreTrainingStratediesForFaceRepresentationLearning}. The other, \cite{SelfSupervised_SpatioTemporal_contrastiveFER},  describes contrastive-learning across the video-sequences, for specifically categorical emotions on acted dataset Oulu-CASIA \cite{Oulu-CASIA}, and is SOTA for acted AERR from video. Additionally, \cite{Kollias_multitask_RECOLA_Affwild} offered a unified framework for multiple tasks, but it does not surpass \cite{RECOLA_VisualResNet50} and \cite{RECOLA_VideoOnly} on RECOLA \cite{RECOLA}.

To the best of our knowledge video-only self-supervised framework for natural apparent emotional reaction recognition has not yet been explored, which is what we present in this paper. We compare 3 different SSL methods for the pretext training and investigate the impact of a variety of loss functions during downstream training. We evaluate our proposed method on \cut{3}two different natural emotional reactions datasets (SEWA\cut{, MSP Face} and RECOLA, \cite{SEWA, RECOLA}) and achieve an improvement by up to 10\% on previously published models.


\textbf{Our main contributions} can be summarized as follows: (1) a review of several pretext tasks for apparent emotional reaction recognition from video for their downstream performance across several \textit{spontaneous} emotion datasets; (2) analysis of the impact of the combined regression and classification losses, data augmentations, and downstream learning parameters; (3) adding up to the first to our knowledge Self-Supervised Visual Apparent Emotional Reaction Recognition method for spontaneous emotions with continuous annotations, SS-VAERR. 
Please check Tab.~\ref{tab:competition} for the results\cut{ and comparisons with other methods}.

\begin{figure*}
    \centering
    \vspace{-5pt}
    \includegraphics[scale=0.23]{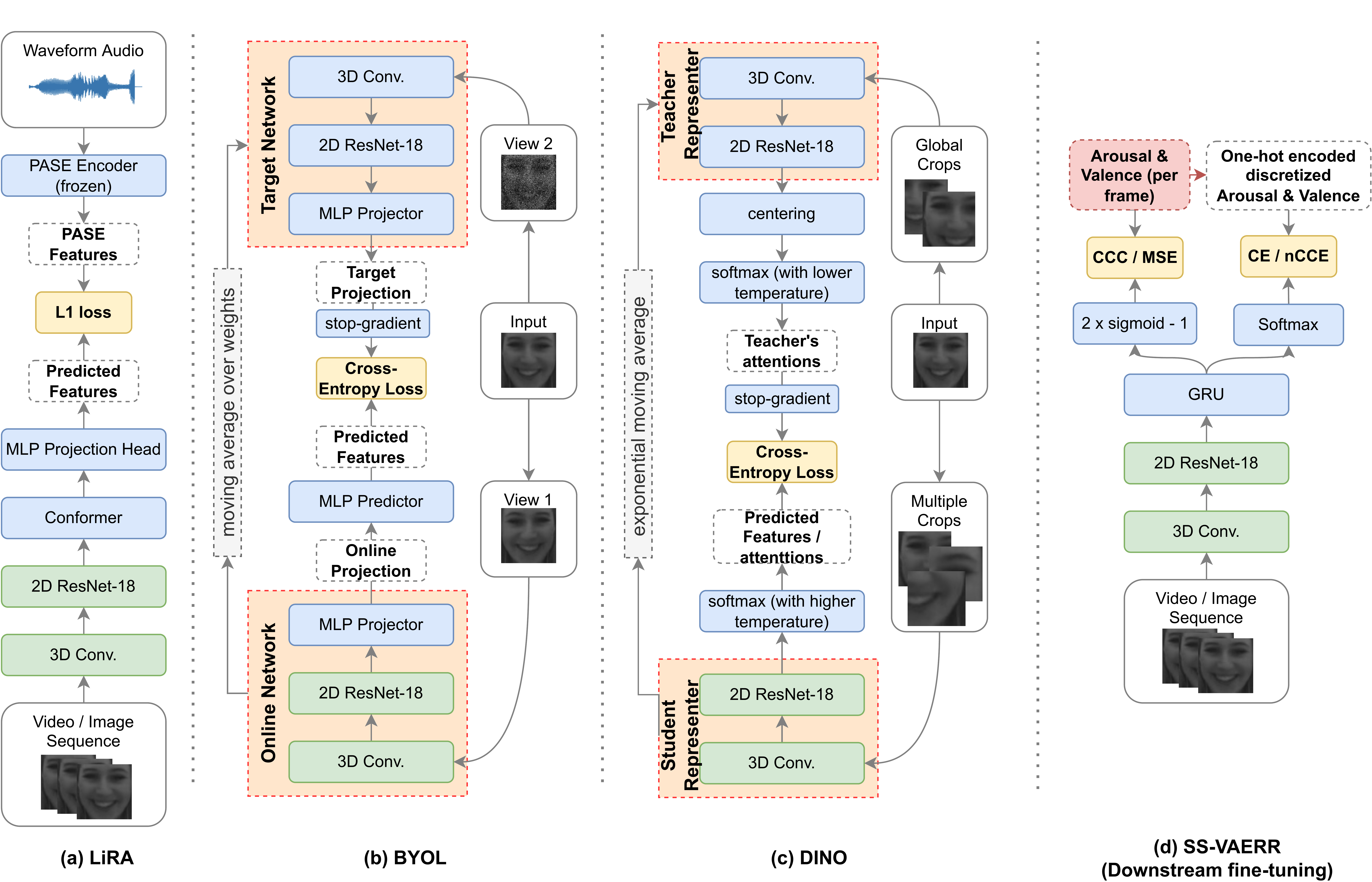}
    \caption{A comparative overview of all the reviewed pretext architectures. Please note 
    ResNet18 architecture is used instead of the original ResNet50 and transformers for BYOL \cite{BYOL} and DINO \cite{DINO} correspondingly for comparability. Blue and green boxes represent the network blocks (layers, standard structures, activations/normalization). Green are the blocks used to initialize the downstream architecture. Yellow boxes represent training losses, and orange are networks with interconnected weights. 
    }
    \label{fig:architectures}
    \vspace{-5pt}
\end{figure*}

\section{Related Work}\label{sec:self-sup}
\textbf{Apparent Emotional Reaction Recognition} is a vast research field spread across different methods and domains. Domain-wise there is audio-based, image-based, video-based, and audio-visual AERR, additionally separated into acted and spontaneous/natural AER. We mostly focus on spontaneous and visual AER here. The results across the field are reported on different datasets, complicating the comparison, that is why SOTA is reported per dataset.
There are also multiple datasets for AERR of different modalities
, the ones discussed in this paper (Sec.~\ref{sec:datasets}), and others, such as AffNet \cite{AffWild_Kollias}, Oulu-CASIA \cite{Oulu-CASIA}, and AffectNet \cite{AffectNet}.

\textit{Audio-visual or multi-modal AERR} tends to yield better results than video-only. Specifically audio is known to provide better signal for arousal \cite{ArousalFromAudio_ValenceFromVideo_review, ArousalFromAudio_ValenceFromVideo_AV+EC, ArousalFromAudio_ValenceFromVideo_AVECWorkshop}.
Multi-modal AERR works include \cite{AlbadawyKim_JointDiscreteAndContinuousEmotionPrediction}, a BLSTM-based method, using joint discrete and continuous emotion representation for AERR that holds the current SOTA for multi-modal AERR on RECOLA \cite{RECOLA} development set. However, a ResNet50-based method presented by \cite{RECOLA_VisualResNet50} achieves SOTA results for RECOLA on the test set, also presenting video- and audio-only results.\cut{Additionally, \cite{MSP_Face_corpus} introduces the MSP Face database, as well as some SOTA results on MSP Face including audio-only, video-only, and audio-visual AERR, these results have been obtained in a complex multiple-level SSL setting.}

The most prominent examples of the \textit{image-based AERR} include EmoFan \cite{Toisol_EstimationOfContinuousValArousalLevelsFromFaces} - an approach for direct estimation of facial landmarks, discrete and continuous emotions with a single neural net from facial images, and \cite{Bulat_PreTrainingStratediesForFaceRepresentationLearning} - an SSL framework, purposed for a variety of downstream face-related applications, including the SOTA results for AERR on images presented on AffectNet \cite{AffectNet}. 

\textit{Video-only AERR} is a less explored field, the current SOTA being Affective Processes \cite{AffectiveProcessess} and \cite{RECOLA_VideoOnly}. First is a neural processes model with a global stochastic contextual representation, task-aware temporal context modelling, and temporal context selection. Second is a
two-stage spatio-temporal attention temporal convolution network. 


\textbf{Self-supervised learning (SSL)} focuses on minimising the use of human-generated annotations at training time. It is often used to leverage the large amounts of unlabelled data to aid learning on significantly smaller annotated datasets.
The \textit{SSL} is rooted in the assumption that solving a seemingly unrelated self-supervised \textit{pretext task} can help to learn useful visual representations. These would serve as a good initialization point for a task of interest, \textit{downstream task}, given that the model is well generalized and the tasks are similar enough in kind \cite{VisualRepresentationsForDwonstream_Mid-levelRepsCNN_Oquab, VisualRepresentationsForDwonstream_VeryDeepCNNForLarge-ScaleRecognition_Simonyan, VisualRepresentationsForDwonstream_RichFeatureHierarchies, VisualRepresentationsForDwonstream_FullyCNNforSemanticSegmentation}. If these assumptions are violated a \textit{negative transfer} might occur \cite{Survey_OnTL} - the performance would be worse than that of a model trained entirely from scratch.

There is plenty of research showing the benefits of SSL for general image datasets \cite{Survey_OnSelfSupervised, Survey_SelfSupervisedRepLearning}. SSL techniques vary by both the downstream and pretext tasks. Traditional pretext tasks include transformation classification \cite{selfSup_rotations_14}, image inpainting \cite{selfSup_inpaining_29}, image colorization (from grayscale) \cite{selfSup_colorization_43, selfSup_colorization_sketch}, and solving jigsaw puzzles \cite{selfSup_jigsaw_26}. A more recent field of SSL is \textit{contrastive learning}, that relies on minimizing the distance between the learned representations of the \textit{positive pairs} - differently augmented versions of the same image, and maximising it for the \textit{negative pairs} - different augmentations of different images, \cite{Contrastive_visualReps_Hinton, Contrastive_momentumContrast_Girshtick}. On the plus side contrastive learning does not require labels, on the downside it is sensitive to the choice of the negative pairs as well as to the choice of the augmentations used.

An ultimate improvement on the contrastive techniques involved getting rid of the negative pairs, accomplished by BYOL\cite{BYOL} and DINO \cite{DINO}. Instead those rely on different teacher-student-like architectures trained on a variety of image crops and augmentations, that can be interpreted as positive pairs. Not relying on negative examples accounts for a potentially better generalization because of their lower vulnerability to the systematic biases in training data. 

\textbf{For the face imagery} some SSL techniques were developed for tasks such as action unit detection \cite{selfSup_AUdetection} and lip-reading \cite{LiRA, selfSup_HUBERT}. For AERR, SSL research can be sparse depending on the modality and target (i.e. categorical emotion vs valence and arousal). The prominent works include \cite{Bulat_PreTrainingStratediesForFaceRepresentationLearning} - universal facial representation learning for images, \cite{SelfSupervisedMultiviewFER} - contrastive learning method for recognition of categorical emotions from multi-view images of emoted faces. 

The only current example of the self-supervised AERR from video is based on spatio-temporal contrastive learning and delivers SOTA results for synthetic categorical emotion recognition \cite{SelfSupervised_SpatioTemporal_contrastiveFER}. Unfortunately, it is not directly comparable to ours because 1) it is designed to perform on a lab recorded and most importantly acted dataset Oulu-CASIA \cite{Oulu-CASIA}; 2) it aims to predict discrete emotions rather than arousal and valence. Meaning not only that it uses non-realistic posed emotion depictions, but also that every discrete acted emotion is present for every individual in the dataset. Whereas we aim for the spontaneous apparent emotional reactions with their natural distribution.
\cut{, with all the natural biases of such setting, e.g., certain emotions being underrepresented for certain individuals, or underrepresented across the entire datasets in general.}

Although some work has been done in multi-modal self-supervised in-the-wild AERR \cite{AlbadawyKim_JointDiscreteAndContinuousEmotionPrediction}, to our knowledge video-only self-supervised continuous spontaneous facial AERR has not been explored yet.

\section{Methodology}\label{sec:methodology}
\subsection{Shared Architecture and Pretext Tasks}\label{sec:pretexts}
As mentioned before, the similarity of the pretext and downstream tasks can significantly contribute to the usefulness of the visual representations learned during pretext training to the downstream task, hence in this study we review and evaluate the self-supervised pretext methods designed to learn visual representations.

We examine three suitable pretext methods: LiRA \cite{LiRA}, BYOL\cite{BYOL}, and DINO \cite{DINO}. Please refer to Fig. \ref{fig:architectures} for the flowcharts complementing the explanations below.
For comparability sake we have incorporated the ResNet18, \cite{ResNet}, as part of the pretext architectures to facilitate the transfer for the downstream task. LiRA architecture, Fig.~\ref{fig:architectures}(a), remains unchanged. For BYOL it is a change from ResNet50 to ResNet18, Fig.~\ref{fig:architectures}(b). For DINO it is a change from transformer, \cite{ViT}, to ResNet18, Fig.~\ref{fig:architectures}(c); in principle DINO supports any architecture for both of its networks.

\textbf{Learning visual speech Representations from Audio (LiRA) \cite{LiRA}} is a self-supervised method for predicting visual representations of acoustic features from unlabelled speech videos. Its architecture features a ResNet18 followed by a conformer. Once it is trained, we only use the ResNet18 weights for the downstream initialization. 
LiRA uses the random flip ($Prob=50\%$) and the random crop of $80\times80$ (out of $96\times96$) augmentations during training.

\textbf{Bootstrap Your Own Latent (BYOL) \cite{BYOL}} is an approach to self-supervised visual representation learning that does not rely on negative pairs typical for previous SSL methods. Its architecture consists of online and target networks, interacting and learning from one another. The online network is trained to predict the invariant visual representation of the same image under different augmentations, while the target network learns via a slow-moving average from the online network. At inference time only the 3D convolutional layer and the ResNet18 of the online network are preserved, the visual representations are extracted from the final layer of ResNet for the downstream task. 
BYOL uses the following training time data augmentations: random cropping, random flip, color jittering (brightness, contrast, saturation and hue of an image, shifted by a uniformly random
offset), Gaussian blurring, and solarization. For details see Appendix B in \cite{BYOL}.


\textbf{Self-DIstillation with NO labels (DINO) \cite{DINO}} is yet another two-network SSL architecture, an attention-based self-distillation method using no labels, and reportedly an improvement on BYOL. The principle is also rather similar to BYOL: the two networks are student and teacher networks, they have the exact same architecture and teacher is updated as an exponential moving average of the student. There are certain additional tricks, such as centering and sharpening for the teacher network. Sharpening is a technique introducing the temperature parameter into the softmax of both networks. Temperature is lower for the teacher than for the student, it reduces noise but encourages a potential \textit{mode collapse} (a phenomenon where network systematically produces same outputs for different inputs), whereas centering (a type of normalization specific to DINO technique with respect to teacher's previous logits) is meant to compensate for that and prevent the mode collapse.
In principle DINO allows for both student and teacher to draw from a broad range of potential network architectures. We have adapted it to rely on the ResNet18 instead of the typical transformer networks for comparability with the other reviewed pretext models. 
DINO uses the same augmentations as BYOL
. However, one of the key differences of DINO is that the teacher network only sees the global crops (covering most of the image) while the student gets to see both global and local crops and should derive similar or ideally the same representations from both.

Please note that the data augmentations during the pretext training are as in the corresponding, whereas the ones at the downstream training time are discussed in Sec.~\ref{sec:results_augmentations}.

\subsection{Downstream VAERR Architecture} \label{sec:architecture}
As can be seen on Figure~\ref{fig:architectures}(d), the downstream task architecture uses 3D convolutional layer, followed by ResNet18, and finally by GRU, which then returns regression or classification result or both depending on what loss is chosen. For all of the pretext tasks the weights of the 3D convolutional layer and ResNet18 are passed from the pretext to downstream architecture, as the initialisation of the latter. The intuition behind the choice of the architecture is follows: ResNet18 is a reliable choice for initial image processing resulting in meaningful latent features, powered by pre-text tasks, whereas GRU is meant to capture the temporal component within the videos.

\textit{Fine-tuned} version of each pretext method only initializes the downstream task with the shared weights from the pretext. It is later free to update these weights in accordance to the downstream training under the chosen loss function. \textit{Frozen} version of the pretext freezes some of the layers, so they cannot be updated during the downstream training of the model. We have compared freezing the entire set of shared weights and found that freezing just the first 3D-convolutional layer of the network generally shows better results that freezing 3D-convolutional layer and ResNet18. Therefore, "Frozen" in this paper means ``frozen first layer''.

\subsection{AERR Discrete and Continuous Labels}\label{sec:labels&losses}
Humans tend to generalize and discretize the facial emotions into 7-8 categrical classes such as happiness, sadness, surprise, anger, rage, disgust, boredom (and neutral). 
From the computational point of view, there are multiple perks to using the continuous emotion labels - valence (positivity/negativity of the emotion) and arousal (the magnitude of it) \cite{Toisol_EstimationOfContinuousValArousalLevelsFromFaces}. Therefore, a number of sentiment analysis datasets are annotated with continuous emotion labels only. In this paper we focus specifically on correctly predicting apparent arousal and valence.

Furthermore, there are several publications in the field suggesting that using combined classification and regression losses, called \textit{composite losses} in this paper, on both continuous and discretized versions of the same labels improves the prediction quality drastically \cite{AffectiveProcessess, AlbadawyKim_JointDiscreteAndContinuousEmotionPrediction}. Hence, we study composite loss functions as well as various pretext tasks.

\setlength{\abovedisplayskip}{3pt}
\setlength{\belowdisplayskip}{3pt}

\subsection{Composite Loses}
The success of the models is assessed via the Concordance Correlation Coefficient \textit{(CCC)} metric calculated on the combined videos of the test set.
Rather than optimizing the performance solely for CCC, we examine the combinations of the following losses.

\textbf{Regression loss} for continuous predictions, is calculated as $1 - CCC$, where CCC is the Concordance Correlation Coefficient of the validation set. The idea behind this metric is to assess the correlation between the predictions and the targets, while also penalising the signals with the different means more. It can be interpreted as a version of Pearson Coefficient weighted towards predictions with higher errors.
\begin{equation} \label{eq:CCC}
   \Scale[0.85]{CCC(Y, \hat{Y}) = 2 \dfrac{ \mathbb{E}(Y-\mu_Y)(\hat{Y}-\mu_{\hat{Y}})}{\sigma_{\hat{Y}} + \sigma_Y + (\mu_{\hat{Y}} + \mu_{Y})^2}}
\end{equation}

where $Y$ are ground truth labels and $\hat{Y}$ are the predicted values, and $\mu$ and $\sigma$ are their mean and variance.

\textbf{Mean Squared Error \textit{(MSE)}} (for continuous labels) shows how close the predicted values are to the target values. We were inspired to use it by the EmoFAN paper \cite{Toisol_EstimationOfContinuousValArousalLevelsFromFaces}, which suggests that optimising with respect to the MSE tends to improve the performance with respect to the CCC as well.
\begin{equation} \label{eq:MSE}
   \Scale[0.9]{MSE(\hat{Y}, Y) =  \mathbb{E}((Y-\hat{Y})^2)}
\end{equation}

\textbf{Cross-Entropy Loss \textit{(CE)}} is a classification loss for the discretized labels, penalising the divergence of the predicted probability from the actual label. 
Discretization was conducted as following: valence and arousal per frame labels have been split into 20 bins/classes each with uniformly distributed bin boundaries (as in \cite{AffectiveProcessess}). These classes then have been presented as one-hot vector labels of length 20 (per frame). 
So the cross-entropy presented as 
\begin{equation} \label{eq:CE}
\Scale[0.9]{{CE}(Y, \hat{Y}) = - \sum\limits_{l=1}^{L} Y_l log({\hat{Y}}_l)}
\end{equation}
for $L$ classes, here 20 by number of the discrete bins, and $\hat{Y}_i$ is computed as a Softmax probability of each class per label per frame.

\textbf{Cost-sensitive cross-entropy loss \textit{(nCCE)}} function \cite{CCE} with the cost norm loss for discretized labels, similar to \cite{AlbadawyKim_JointDiscreteAndContinuousEmotionPrediction}: 
\begin{equation} \label{eq:nCCE}
\Scale[0.9]{{nCCE}(Y, \hat{Y}) = \dfrac{1}{F} \sum\limits_{f=1}^{F} C_{norm}(Y_{f}, \hat{Y}_{f})\sum\limits_{l=1}^{L}{Y_f}^{(l)}\cdot log{\hat{Y}_f}^{(l)}}
\end{equation}

for $f = 1, ..., F$ being the number of frames and a cost norm function $C_{norm}$, inspired by \cite{AlbadawyKim_JointDiscreteAndContinuousEmotionPrediction}, that takes into consideration the spatial relation which helps the stability of the training / fine-tuning:
\begin{equation} \label{eq:C_norm}
\Scale[0.9]{C_{norm}(Y_{f}, \hat{Y}_{f}) = 1 + \left\| \sum\limits_{l=1}^{L}K^{(l)}({Y_f}^{(l)} - {{\hat{Y}}_f}^{(l)}) \right\| _2}
\end{equation}

where $K^{(l)}$ is the centroid of the label $l$ in k-means classification ($k=20$ in our case).
We analyse the effect of these losses and some of their combinations in Sec.~\ref{sec:results_losses}.



The total loss function can be described as
\begin{equation} \label{eq:total_loss}
\Scale[0.9]{L = w_{ccc} \, CCC + w_{mse} \, MSE + w_{ce} \, {CE}+ w_{_{n}cce} \, {nCCE}}
\end{equation}

\setlength{\tabcolsep}{12pt}
\renewcommand{\arraystretch}{1.05}
\begin{table*}[htbp]
\vspace{-5pt}
\caption{Our proposed model vs state-of-the-art. 
RECOLA: results are presented on development set as the test set is not public. For \cite{AffectiveProcessess} (AP+Det.+Att.) stands for Affective Processes with combined latent and deterministic layers with self-attention.}
\vspace{-5pt}
\begin{center}
\begin{tabular}{l | c c | c c}
\hline
\textbf{Methods}&\multicolumn{2}{|c|}{\textbf{SEWA}}&\multicolumn{2}{|c}{\textbf{RECOLA}} \\
\cline{2-5} 
 & \textbf{Arous.}& \textbf{Val.} & \textbf{Arous.}& \textbf{Val.} \\
\hline
HO-CPConv \cite{video-only-sewa-Kossaifi2020FactorizedHC} & 0.520 & 0.750  & &  \\
Affective Processes (AP+Det.+Att.) \cite{AffectiveProcessess} & 0.662 & 0.672 &  &  \\

Affective Processes Best \cite{AffectiveProcessess} & 0.640 & 0.750 &  &  \\


End-to-End Visual ResNet-50 \cite{RECOLA_VisualResNet50} &  &  & 0.371 & 0.637 \\

TS-SATCN \cite{RECOLA_VideoOnly}&  &  & 0.659 & \textbf{0.690} \\

\hline
Baseline: 3Dconv+ResNet18+GRU From Scratch & 0.588 & 0.609 & 0.344 & 0.538 \\
Our SS-VAERR backbone & 0.678 & 0.737 & 0.630 & 0.607\\
Our SS-VAERR (+ augmentations + composite loss) & \textbf{0.713} & \textbf{0.771} & \textbf{0.675} & 0.626\\
\hline
\hline

\end{tabular}
\label{tab:competition}
\end{center}
\vspace{-10pt}
\end{table*}

\section{Experimental Setup}\label{sec:experiments}
\subsection{Datasets and Preprocessing}\label{sec:datasets}
\textbf{Pretext Dataset} In order to maintain the comparability across different pre-training methods in this paper all of the pretext tasks are trained on Lip Reading Sentences 3 dataset (LRS3) \cite{LRS3}, containing thousands of spoken sentences from TED and TEDx videos. 

\textbf{Downstream Datasets} The downstream task results are presented for the following facial video datasets: SEWA \cite{SEWA}\cut{, MSP Face \cite{MSP_Face_corpus}}, and RECOLA \cite{RECOLA}, the most popular academic datasets for AERR. \textit{SEWA} database consists of the videos of volunteers watching adverts chosen to elicit apparent emotional reactions, and later discussing what they have seen. SEWA has annotations of valence and arousal per frame. SEWA dataset has been collected across the residents of 6 countries: the UK, Germany, Hungary, Serbia, Greece, and China. \cut{ \textit{MSP Face} is an audiovisual database obtained from video-sharing websites, where multiple individuals discuss various topics expressing their opinions and experiences. Emotions in MSP Face are annotated as a single label per video. For comparability of the results with the other two datasets we use this single video label as a constant label per frame.} \textit{RECOLA} is a database of multi-domain data recordings of native French-speaking participants completing a collaborative task in pairs during a video conference call, collected in France. Although RECOLA in the wild possesses a rich choice of modalities, we only use the pre-processed video data and continuous arousal and valence labels recorded per frame, averaged across the annotators.

\textbf{Pre-processing} All of the above datasets are converted into gray-scale videos and cropped around the face to $96\times96$ based on the landmark detection. More specifically we use RetinaFace face detector \cite{Retinaface_2020_CVPR} and the Face Alignment Network (FAN) \cite{FAN_CVPR2017} to detect 68 facial landmarks and crop the face based on these. Annotations include arousal and valence, labelled per frame, averaged across multiple annotators\cut{, and extended to per frame labels in case of a single annotation per video (for MSP Face)}.

\setlength{\tabcolsep}{12pt}
\renewcommand{\arraystretch}{1.05}
\begin{table*}[htbp]
\caption{Comparison of the pretext techniques across various datasets for video-only AERR.}
\vspace{-5pt}
\begin{center}
\begin{tabular}{l l | c c | c c }

\hline
& &\multicolumn{2}{|c|}{\textbf{SEWA}}&\multicolumn{2}{|c}{\textbf{RECOLA}} \\
\cline{3-6} 
&\textbf{ } & \textbf{Arous.}& \textbf{Val.} & \textbf{Arous.}& \textbf{Val.} \\
\hline
&+ LIRA frozen & 0.652 & 0.722 & 0.602 & 0.532\\
&+ LIRA fine-tuned & \textbf{0.678} & \textbf{0.737} & 0.630 & \textbf{0.607} \\


\textbf{PRETEXT}&+ Video-BYOL frozen & 0.593 & 0.726 & 0.224 & 0.344\\
\textbf{TECHNIQUES}&+ Video-BYOL fine-tuned & 0.604 & \textbf{0.757} & 0.307 & 0.446\\

&+ DINO-ResNet frozen & 0.607 & 0.638 & 0.269 & 0.545 \\
&+ DINO-ResNet fine-tuned & 0.648 & 0.667 & 0.420 &	0.520 \\

\hline
\hline
\end{tabular}
\label{backbones}
\end{center}
\vspace{-5pt}
\end{table*}

\subsection{Training details}\label{sec:datasets}
Breakdown into training, validation, and test set is conducted in the same manner as in \cite{AffectiveProcessess} for SEWA (train./val./test sets containing 435/53/53 instances)\cut{, as in \cite{MSP_Face_corpus} for MSP Face (10723/1341/2485)}, and as in \cite{AlbadawyKim_JointDiscreteAndContinuousEmotionPrediction} for RECOLA (train./val. containing 197/152 instances, with results reported on the validation set, as the test set for RECOLA is not publicly available). \cut{All}Both of the datasets are normalized with their respective means and standard deviations throughout.

Videos are kept at original lengths and sampled as fixed length segments at training time. Empirically, \cut{MSP Face and }SEWA experiments yield better results with segments of $200$ frames, whereas RECOLA experiments deliver better results when sampled as $500$-frame-long video segments.

During training we have used AdamW optimizer with the weight decay of 0.0001 and initial learning rate ranging from 0.0003 to 0.00007, depending on the downstream dataset and other parameters, and batch-size ranging from 3 to 20. All models in this paper have been trained for 10 epochs. The augmentations are discussed in sections and \ref{tab:augmentations}.


\setlength{\tabcolsep}{4pt}
\renewcommand{\arraystretch}{1.05}
\begin{table*}[htbp]
\caption{Comparison of the various losses for the downstream tasks with LiRA pre-training. Only non-zero loss-weights are presented. `Arous.' and `Val.' superscripts specify the loss applied specifically to either arousal or valence predictions.}
\vspace{-5pt}
\begin{center}
\begin{tabular}{l l | c c | c c | c c | c c }

\hline
& &\multicolumn{4}{|c|}{\textbf{SEWA}}&\multicolumn{4}{|c}{\textbf{RECOLA}} \\
\hline
& &\multicolumn{2}{|c|}{\textbf{Fine-Tuned}}&\multicolumn{2}{|c}{\textbf{Frozen}}&\multicolumn{2}{|c|}{\textbf{Fine-Tuned}}&\multicolumn{2}{|c}{\textbf{Frozen}} \\
\cline{3-10} 
&\textbf{ } & \textbf{Arous.}& \textbf{Val.} & \textbf{Arous.}& \textbf{Val.}& \textbf{Arous.}& \textbf{Val.} & \textbf{Arous.}& \textbf{Val.} \\
\hline

\textbf{REGRESSION} & $w_{ccc}=1$ & 0.678 & \textbf{0.737 }& 0.652 & 0.722 & 0.630 & 0.607 & 0.560 & 0.603 \\
\textbf{LOSSES} & $w_{mse}=1$ & 0.664 & 0.726 & 0.648 & 0.710 & 0.399 & 0.596 & 0.394 & 0.596 \\
\hline

&$w_{ccc}=0.5, w_{ce}=0.5$ & 0.671 & \textbf{0.735} & 0.650 & \textbf{0.747} & 0.454 & 0.625 & 0.513 & 0.606 \\
&$w_{ccc}=0.5, w_{ce}=0.25, w_{mse}=0.25$ & \textbf{0.716} & 0.731
&0.699 & \textbf{0.747} &0.473 & 0.611 & 0.469 & 0.610 \\

\textbf{COMPOSITE} & $w^{Val.}_{ccc}=1, w^{Arous.}_{ccc}=0.66, w^{Arous.}_{ce}=0.34$& 0.631 & 0.663 & 0.659 & 0.709 & \textbf{0.675} & 0.626 & 0.640 & \textbf{0.668} \\
\textbf{LOSSES} & $w^{Val.}_{ccc}=1, w^{Arous.}_{ccc}=0.66, w^{Arous.}_{ce}, w^{Arous.}_{mse}=0.17$& 0.638 & 0.716 & 0.658 & 0.691 & 0.664 & 0.644 & \textbf{0.655} & 0.605 \\

& $w_{ccc}=0.5, w_{_{n}cce}=0.5$ & 0.633 & 0.667 & \textbf{0.701} & 0.741 & 0.669 & 0.655 & 0.614 & 0.661 \\
& $w_{ccc}=0.5, w_{_{n}cce}=0.25, w_{mse}=0.25$& 0.669 & 0.716 & 0.633 & 0.733 & 0.606 & \textbf{0.669} & 0.626 & 0.623 \\

\hline
\hline

\end{tabular}
\label{tab:losses}
\end{center}
\vspace{-10pt}
\end{table*}

\section{Results}\label{sec:results}

\subsection{Comparison with state-of-the-art} \label{sec:baselines}
A slight complication natural to the field is that the results are being presented on a variety of datasets that do not coincide between the papers. Hence, we are restricting our benchmarks by the modality and type of the prediction - video-only natural AERR for valence and arousal. This leaves us with only a few recent benchmarks. First, \cite{AffectiveProcessess} and \cite{video-only-sewa-Kossaifi2020FactorizedHC}, demonstrated their results on SEWA dataset, one of our downstream task datasets. \cut{We also provide the result for \cite{MSP_Face_corpus}, still the best published result for the MSP Face apparent emotional reactions recognition. }For RECOLA there is TS-SATCN \cite{RECOLA_VideoOnly}, a two-stage spatio-temporal attention temporal convolution network, and a visual ResNet50 \cite{RECOLA_VisualResNet50}. That is it for the spontaneous video-only AERR for arousal and valence. 

Since these benchmarks were not evaluated on the same dataset, we compare our results in Table \ref{tab:competition} to their reported results on the respective datasets: SEWA for \cite{AffectiveProcessess, video-only-sewa-Kossaifi2020FactorizedHC}, and RECOLA for \cite{RECOLA_VisualResNet50, RECOLA_VideoOnly}\cut{, and MSP Face for \cite{MSP_Face_corpus}}. Our final model is pre-trained in a self-supervised manner, using augmentations and composite losses (a detailed analysis for each of them can be found in sections \ref{sec:results_pretexts} to \ref{sec:results_augmentations}). We see that our model  compares favourably to the reported results for most of these, confidently outrunning \cite{video-only-sewa-Kossaifi2020FactorizedHC, AffectiveProcessess}. It also outperforms\cut{ \cite{MSP_Face_corpus},} \cite{RECOLA_VideoOnly} and the visual network from \cite{RECOLA_VisualResNet50} for arousal, and only behind these a little (but still at a comparable level) for valence.

\textbf{Baseline} For completeness we present the results for our model stripped of the self-supervised component, trained from scratch, called 3Dconv+ResNet18+GRU in Table~\ref{tab:competition}. 

\subsection{Empirical comparison of pretext tasks}\label{sec:results_pretexts}

First, we compare the performance of different pretext methods - LiRA, BYOL, and DINO used to pre-train 3D convolutional layer + ResNet18 on LRS3 and then fine-tune and assess the performance of the downstream architecture (3D convolutional layer + ResNet18 + GRU) in terms of the CCC across video-only facial datasets. Please see Table~\ref{backbones} for the results. 
Please note that the pretext techniques only have the basic CCC-based regression loss at this point.

It appears that LiRA pretext initialization fine-tuned on the downstream task seem to perform generally better than the other pretext methods. It achieves either the best or the second-best results across all the datasets for both arousal and valence. It even beats some of the benchmarks, despite not yet benefiting from the composite losses used by these methods. The DINO-ResNet18 also delivers results comparable to the benchmarks on most datasets.

There are several potential reasons why LiRA performs better than DINO and BYOL. First of all LiRA uses a temporal model, while others are trained per frame, which might affect the quality of learnt representations for video. It also uses the audio input for guided learning of the visual representations, the other two do not, which might help learning more emotion-relevant representations. Finally, DINO and BYOL usually rely on larger networks trained the datasets of scale that simply not available in AERR field.

The results of this section certainly support the hypothesis that the self-supervised pre-training can be highly beneficial in the video-only spontaneous AERR scenarios.

\subsection{Empirical Comparison of Training Losses}\label{sec:results_losses}
In this section we explore the impact of the various auxiliary loss functions on the performance of the pretext method LiRA during the downstream fine-tuning. We focus on LiRA pretext from now on since it has been identified as the best pretext for our purposes in the previous section.

Please refer to the Table~\ref{tab:losses} for the results on the set of experiments concerning the downstream loss functions. The first line corresponds to the loss used during the pretext task analysis in Table~\ref{backbones}. The benchmarks are the same as in the Table~\ref{backbones} as well, except now the comparison to them is more fair as the presented results include the loss design. The benchmarks are relevant for both fine-tuned and frozen weights versions, however we enter them in the fine-tuned section since they themselves lack this distinction in their design, so closer to fine-tuned in fashion.

Evidently, adding the CE classification loss (Eq.~\ref{eq:CE}) on SEWA improves the performance for apparent arousal, whereas valence seem to either benefit minimally or show worse results. Further adding MSE (Eq.~\ref{eq:MSE}) achieves state-of-the-art results on SEWA.
For RECOLA neither CE classification loss nor MSE loss on their own seem to have a positive impact, however together they create a minor performance boost for valence. Further adjusting to only penalising the apparent arousal loss with CE and MSE leads to a considerable boost in arousal as well. These final results outperform the current video-only SOTA for estimating arousal, while lacking a little for estimating valence \cite{RECOLA_VideoOnly}.

The reason why estimating the apparent arousal might require some more careful loss design in this case is not entirely clear. However, it can be in part explained by the fact that apparent arousal tends to be more present and easier detected in audio, rather than in video, while valence tends to exhibit the opposite trend \cite{ArousalFromAudio_ValenceFromVideo_review, ArousalFromAudio_ValenceFromVideo_AV+EC, ArousalFromAudio_ValenceFromVideo_AVECWorkshop}. Given that we are restricted to the video-only modality it is only natural that achieving good results on arousal requires more parametrization than apparent valence.

Using nCCE (Eq.~\ref{eq:nCCE}) instead of CE often yields close second or occasional best results, it can be viewed as a solid alternative to CE.

\cut{Additionally, we found that MSP Face does not seem to improve from the composite loses in general, we assume that this is because of the different format of the annotations compared to the other datasets. Since MSP Face is annotated per video unlike SEWA and RECOLA that are labeled per video frame, we extended MSP Face labelling to per frame. This might lead to an insufficiently granular correlation between the change in frames vs their labels for a relatively subtle  mechanism of auxiliary losses to have an effect.}

\cut{Aside from these irregularities, t}The hypothesis postulated earlier, as well as in \cite{AlbadawyKim_JointDiscreteAndContinuousEmotionPrediction} and \cite{AffectiveProcessess}, holds. Confirming that the better performing models tend to use combinations of various regression and classification losses, i.e. Equation ~\ref{eq:total_loss} with various weight parameters, resulting in improvements over the classic CCC-based loss function (Eq.~\ref{eq:CCC}).

\subsection{The Impact of the Data Augmentation}\label{sec:results_augmentations}
\begin{figure}
\begin{tabular}{c c c c c c}
 Orig. & Hor.Flip & Rand.Cr. & CutOut & Solar. & S\&P\\
 
\includegraphics[width=13mm]{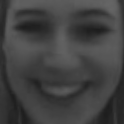} 
&
\includegraphics[width=13mm]{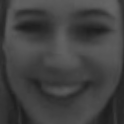}\vspace{0.5pt} & \includegraphics[width=13mm]{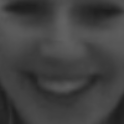}\vspace{0.5pt} & \includegraphics[width=13mm]{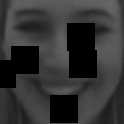}\vspace{0.5pt}& \includegraphics[width=13mm]{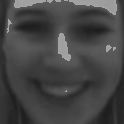}\vspace{2pt} & \includegraphics[width=13mm]{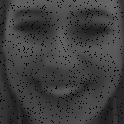}\\

\end{tabular}
\caption{Examples of augmentations used in Sec.~\ref{sec:results_augmentations}.}
\label{fig:augs}
\vspace{5pt}
\end{figure}

Augmentations for the pretext tasks are preserved as in their corresponding publications \cite{LiRA, BYOL, DINO}. 
The experimental results presented up until this point are performed without any augmentations during the downstream training. In this section we present the ablation for different types of augmentations applied during the downstream training (e.g. Fig.~\ref{fig:augs}). We conduct these experiments on the best performing models for their respective datasets:
\begin{itemize}
    \item SEWA: LiRA pretext with composite loss of $w_{ccc}=0.5, w_{ce}=0.25, w_{mse}=0.25$ fine-tuned;
    \item RECOLA: LiRA pretext with composite loss of $w^{Val.}_{ccc}=1, w^{Arous.}_{ccc}=0.66, w^{Arous.}_{ce}=0.34$ fine-tuned.
\end{itemize}

\setlength{\tabcolsep}{1pt}

\setlength{\tabcolsep}{3.5pt}
\renewcommand{\arraystretch}{1.05}
\begin{table}[htbp]
\caption{Downstream Data Augmentation Ablation Results}
\vspace{-10pt}
\begin{center}
\begin{tabular}{l | c c | c c }
\hline
\textbf{LiRA}&\multicolumn{2}{|c|}{\textbf{SEWA}}&\multicolumn{2}{|c}{\textbf{RECOLA}} \\
\cline{2-5} 
\textbf{with Composite Loss} & \textbf{Arous.}& \textbf{Val.} & \textbf{Arous.}& \textbf{Val.} \\
\hline
No Augmentations & 0.746 & 0.747 & 0.675 & \textbf{0.626}\\
\hline
\hline
Horizontal Flip ($Prob = 0.5$) & 0.713 & \textbf{0.771} & 0.636 & 0.561\\
Random Crop & 0.629 & 0.738 & 0.611 & 0.490 \\

Crop-Out 25\% & 0.688 & 0.734 &  0.612 & 0.563 \\
Missing Frames 20\% & 0.681 & 0.750 &0.545 & 0.587 \\
Solarization 20\% & 0.680 & 0.741 & 0.149 & 0.028 \\
Salt \& Pepper Noise & 0.678 & 0.727 & 0.346 & 0.553 \\
\hline
\hline
Rand.Crop + Horiz. Flip& 0.637 & 0.653 & 0.601 & 0.395\\
Rand.Crop + Missing Frames& 0.617 & 0.702 & 0.506 & 0.373 \\
Horiz. Flip + Missing Frames& 0.655 & 0.761 & 0.649 & 0.004\\
Horiz. Flip + Solarization& 0.679 & 0.721 & 0.024 & 0.037 \\
Horiz. Flip + Crop-Out& 0.694 & 0.685 & \textbf{0.701} & 0.451 \\
Horiz. Flip + S\&P& \textbf{0.716} &0.761 & 0.688 & 0.445 \\

\hline
\hline
\end{tabular}
\label{tab:augmentations}
\end{center}
\vspace{-10pt}
\end{table}

For the detailed results please refer to Table~\ref{tab:augmentations}. This list of the augmentations is a compilation of the LiRA and BYOL-recommended sets of augmentations used at the pretext training time, with some minor exceptions such as the colour-related augmentations, since we use the grayscale versions of the downstream datasets. We also explore promising combinations of the individual augmentations.

\textit{Horizontal Flip} is a mirror reflection of an image, applied with 50\% probability to training images.
\textit{Random Crop} is cropping an image to the smaller size at random (size being $110\times110$ for SEWA and $80\times80$ for RECOLA) for all training images.
\textit{Crop-Out} occludes several patches in the image (in our case 5 patches of square shape) with black boxes.
\textit{Missing Frames} means 20\% of frames at random replaced by black frames at training time. \textit{Solarization} is a phenomenon in photo-imaging where the image is wholly or partially reversed in tone. 
In this case solarization refers to an unnatural lighting effect, like in figure~\ref{fig:augs}. Is not as popular in data augmentation techniques, however \cite{BYOL} found it beneficial for their model. We solarize above the average lightness, the effect is applied to 20\% of the training images.
\textit{Salt \& Pepper Noise} - classic noise with 50\% salt vs pepper split, applied to all training images.

The best results on SEWA datasets are provided by the horizontal flip and its combination with the salt \& pepper noise. The rest of the augmentations do not seem to bring any significant improvement and, in fact, often worsen the performance. For RECOLA augmentations seem to almost always have a negative effect on the performance. Nevertheless, there is a specific instance for the horizontal flip and crop out combination where the performance for arousal gets close to even some of the multi-modal results \cite{RECOLA_VisualResNet50}.

\section{Discussion and Conclusion}\label{sec:discussion}

To conclude, in this paper we have presented the first to our knowledge a self-supervised technique for the video-only natural apparent emotional reactions recognition, yielding the current state-of-the-art (or closely comparable) results for video-only natural AERR. Complete with comparative empirical study of the potential pretext methods, auxiliary loss functions, and downstream-time data augmentation ablation. We also found that the optimal parameter search is somewhat unsurprisingly data-dependent, whereas the self-supervised setting is on average beneficial.

Additionally, we argue that the facial apparent emotional reactions recognition is highly data-specific. Factors that should be considered can include: the source and distributions of the pretext and downstream data (acted vs spontaneous, lab-recorded vs in-the-wild, outdoors vs indoors, speaking vs passive listening faces), as well as the specific data preprocessing procedures, format and configuration of the annotations (per frame vs per video, categorical vs continuous emotion annotation), etc.

Previous research suggests that different modalities tend to provide better cues for different apparent emotion metrics: video tends to be a better indicator for the video-aided recognition, and arousal tends to be better detected from audio modality \cite{ArousalFromAudio_ValenceFromVideo_review, ArousalFromAudio_ValenceFromVideo_AV+EC, ArousalFromAudio_ValenceFromVideo_AVECWorkshop}. This makes the video-only AERR particularly challenging in terms of identifying the correct levels of arousal, and explains valence-arousal discrepancy for several results in this paper.

We present the results confirming that using the self-supervised setting alone helps beating (or at least reaching comparable results with) the current state-of-the-art without even touching upon the loss function design. Next we present the evidence that the careful composite loss design can further improve the performance. And finally we provide an ablation on potentially beneficial data augmentation techniques which can lead to further improvements.

Future work could be extended to a comparative analysis of the impact of the different pretext datasets, along with the pretext training parameters and data augmentations. Additional study could be conducted on the specifics of the downstream architectures, as well as investigating the effect of sharing the learned feature representations including temporal component (for architectures with such a component) in order to fully exploit the potential of the video domain.

\section{Acknowledgements}
We would like to thank Pingchuan Ma for for providing the pre-processing and LiRa pre-training code.



{\small
\bibliographystyle{IEEEtran}
\bibliography{IEEEexample.bib}
}

\end{document}